\pdfoutput=1

\documentclass[11pt]{article}
\usepackage[final]{coling}

\usepackage{times}
\usepackage{latexsym}
\usepackage[T1]{fontenc}
\usepackage[utf8]{inputenc}
\usepackage{microtype}
\usepackage{inconsolata}
\usepackage{graphicx}
\usepackage{hyperref}
\usepackage{kotex}
\usepackage{graphicx}
\usepackage{amssymb}
\usepackage{pifont}

\newcommand{\cmark}{\ding{51}}%
\newcommand{\xmark}{\ding{55}}%
\usepackage{booktabs}
\usepackage{enumitem}
\setitemize{noitemsep,topsep=2pt,parsep=0pt,partopsep=0pt}
\usepackage{threeparttable}
\usepackage{multirow}
\usepackage{adjustbox}
\usepackage{colortbl}
\usepackage{blindtext}
\usepackage{enumitem}
\usepackage{url} 
\usepackage{tocloft}

\title{VLR-Bench: Multilingual Benchmark Dataset for Vision-Language
Retrieval Augmented Generation}

\author{
    \textbf{Hyeonseok Lim\thanks{These authors contributed equally.}, Dongjae Shin$^{\ddagger*}$, Seohyun Song, Inho Won$^{\ddagger}$}\\
    \textbf{Minjun Kim, Junghun Yuk$^{**}$, Haneol Jang$^{**\dagger}$, KyungTae Lim\thanks{Corresponding authors.}} \\ 
    Seoul National University of Science and Technology (SeoulTech)\\
    $^{\ddagger}$SeoulTech \& Teddysum \\
    $^{**}$Hanbat National University \\
    \texttt{\{gustjrantk,dylan1998,alexalex225225,wih1226,mjkmain\}@seoultech.ac.kr,} \\
    \texttt{20191780@hanbat.ac.kr, hejang@hanbat.ac.kr, ktlim@seoultech.ac.kr}
}

\begin{document}
\maketitle

\addtocontents{toc}{\protect\setcounter{tocdepth}{0}}
\begin{abstract}
We propose the \textsc{VLR-Bench}, a visual question answering (VQA) benchmark for evaluating vision language models (VLMs) based on retrieval augmented generation (RAG). Unlike existing evaluation datasets for external knowledge-based VQA, the proposed \textsc{VLR-Bench} includes five input passages. This allows testing of the ability to determine which passage is useful for answering a given query, a capability lacking in previous research. In this context, we constructed a dataset of 32,000 automatically generated instruction-following examples, which we denote as VLR-IF. This dataset is specifically designed to enhance the RAG capabilities of VLMs by enabling them to learn how to generate appropriate answers based on input passages. We evaluated the validity of the proposed benchmark and training data and verified its performance using the state-of-the-art Llama3-based VLM, the Llava-Llama-3 model. The proposed \textsc{VLR-Bench}\footnote{\scriptsize https://huggingface.co/datasets/MLP-KTLim/VLR-Bench} and \textsc{VLR-IF}\footnote{\scriptsize https://huggingface.co/datasets/MLP-KTLim/VLR-IF} datasets are publicly available online.
\end{abstract}
\section{Introduction}
\label{sec:Introduction}
The search for external knowledge is very important for VLMs because it is often impossible to find answers directly from images in response to user queries \citep{ok-vqa}.
Previous studies attempted to incorporate external knowledge into VLMs. Among these efforts, dense passage retrieval~\citep{karpukhin2020dense} has been used to search for documents related to queries in an attempt to solve this problem \citep{luo2021weakly,gao2022transform}. However, as \citet{lin2022retrieval} pointed out, these models face challenges in determining whether the retrieved documents are useful for answering queries.Following this, the proposed RA-VQA \citep{lin2022retrieval} introduced an approach that simultaneously conducts searches and question-answering to overcome these drawbacks. However, since the study primarily focused on the RAG configuration, evaluating how the VLM utilized the search results remained challenging.

To address these issues, we propose a Vision Language-RAG Benchmark (\textsc{VLR-Bench}) and training data to evaluate the Retrieval-Augmented Generation (RAG) capabilities of VLMs \citep{rag}. \textsc{VLR-Bench} consists of 300 datasets composed of problems that are difficult to solve without external knowledge. The data were structured as an image-query-passage-output, and unlike conventional VQA datasets, each dataset contained five distinct passages. Only two passages contained direct information that could resolve the queries. This allows us to test the ability, which has been lacking in previous research, to determine which passages are useful for answering queries. \setlength{\parskip}{0em} 

In this study, we developed the VLR Instruction Following (\textsc{VLR-IF}) training data for VLM RAG based on the data generation method proposed by LLaVA \citep{llava} and assessed its utility.
We validated the proposed \textsc{VLR-Bench} and \textsc{VLR-IF} training data based on the following three research questions: (1) Does the proposed \textsc{VLR-Bench} require external knowledge retrieval to be solved? (2) How does the proposed training data impact external knowledge utilization? (3) How effectively can public VLMs and commercial models resolve queries that require retrieval? \setlength{\parskip}{0em} 

In this study, we conducted a baseline performance evaluation of \textsc{VLR-Bench} using the most recently released vision language models in the \textsc{Llava-Llama-3} series \citep{2023xtuner} and GPT-4o \citep{openai2024gpt4}. The contributions of this study can be summarized as follows: 
\begin{itemize}
\item We propose multilingual RAG evaluation data, \textsc{VLR-Bench}, and training data, \textsc{VLR-IF}, for the VLMs.
\item Through in-depth analysis, we prove the actual effect of our dataset.
\end{itemize}
\section{Related Work}
\label{sec:Related Work}
\paragraph{VLM Benchmark Datasets.}
In the VLM benchmark, OK-VQA \citep{ok-vqa} is a key open-domain VQA dataset that uses external knowledge from Wikipedia. Subsequently, A-OKVQA \citep{a-okvqa} and S3VQA \citep{s3vqa}, which included justifications for answers, were derived from OK-VQA. Additionally, datasets targeting specific domains have appeared; for instance, K-VQA \citep{k-vqa}, which intensively utilizes personal information, and ViQuAE \citep{viquae}, which uses object information, were proposed as evaluation datasets. Furthermore, VQA models utilizing knowledge graphs have been proposed, notably GQA~\citep{Hudson_2019_CVPR}, which uses scene graph knowledge and its multilingual expansion (xGQA~\citep{pfeiffer-etal-2022-xgqa} and BOK-VQA~\citep{kim2024bokvqa}). In a different context, datasets providing passages for evaluating the RAG capabilities of VLMs have recently emerged. 
Notable examples include InfoSeek~\citep{infoseek} and Encyclopedic VQA~\citep{encyclopedic}. These datasets provide passages or entire documents, resulting in performance variations based on the document retrieval ability. Detailed information on these external knowledge-based VLM benchmark datasets, as well as their differences from the proposed \textsc{VLR-Bench}, can be found in Appendix~\ref{sec:appendix}.
\section{Proposed RAG Dataset for VLMs} 
\label{sec:dataset}
Benchmarks related to the use of external knowledge by VLMs, as discussed in Section~\ref{sec:Related Work}, particularly InfoSeek and Encyclopedic VQA, typically provide single gold-standard evidence to resolve queries. However, real-world RAG-based systems generate answers by incorporating multiple retrieved results (e.g., Top-5). A significant challenge arises when plausible but incorrect information is retrieved as external knowledge. Therefore, when VLM models use RAG, it is essential to evaluate (1) how accurately external knowledge is retrieved and (2) the model's ability to generate correct answers despite the existence of incorrect information. In this context, we propose \textsc{VLR-Bench}, which simultaneously considers the correct selection of external knowledge and answers-generated by VLMs. In addition, we introduce a construction method for the \textsc{VLR-IF} dataset designed to enhance the ability of VLMs to select external knowledge.

\subsection{VLR-Bench Dataset}
\label{sec:VLR-Bench}
\begin{figure}
  \includegraphics[width=\linewidth]{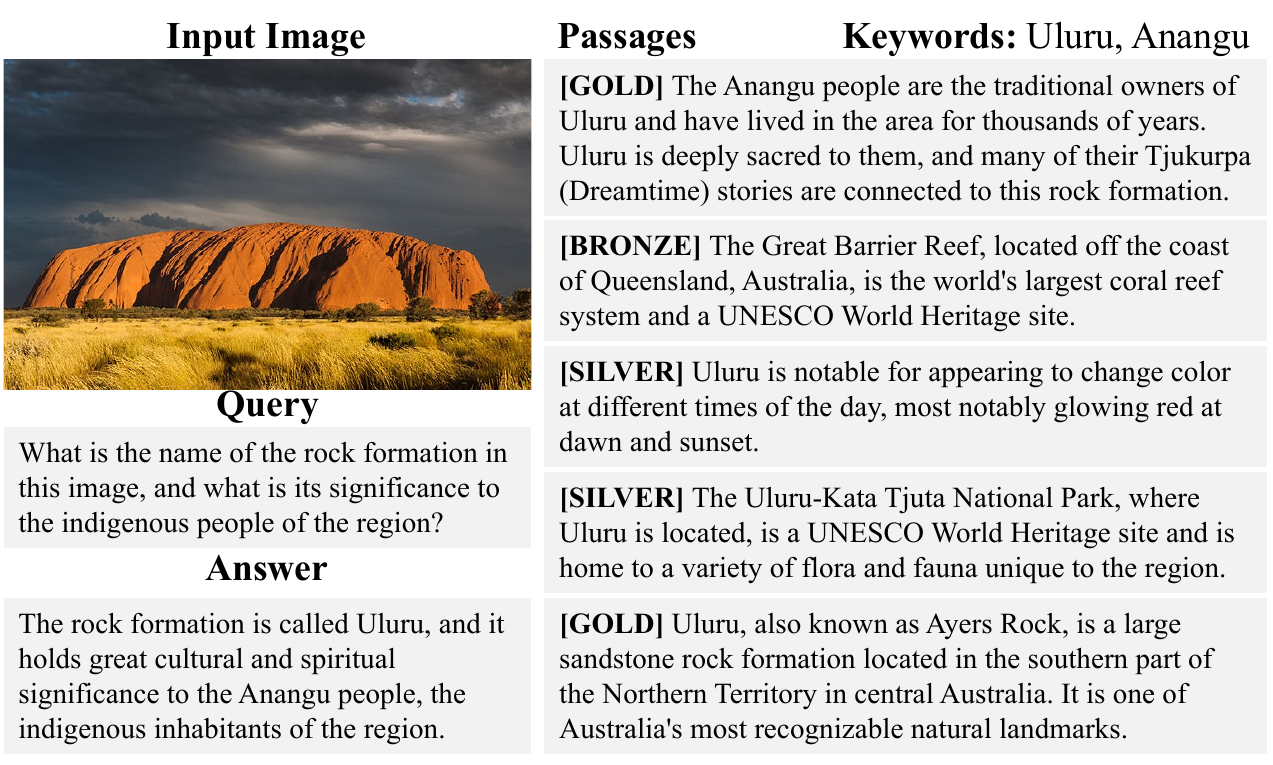}
  \captionsetup{font=small}
  \caption{An example of \textsc{VLR-Bench} data sample.}
  \label{fig:V-RAG data exam}
\end{figure}
\textsc{VLR-Bench} was constructed to evaluate whether VLMs can use the correct external knowledge to generate accurate responses to query. We constructed a parallel corpus of 300 datasets: 150 based on general knowledge and 150 based on cultural data from English, Chinese, and Korean. Detailed examples of the data are provided in Appendix~\ref{appendix:vlr-bench}. 

\begin{table*}[!h]
\tiny
\centering
\begin{tabular}{llccccccccccccc}
\toprule

\multirow{2}{*}{Lang.} & \multirow{2}{*}{Model} &\multicolumn{3}{c}{VLR-IF} & \multicolumn{5}{c}{With Passages} & \multicolumn{5}{c}{Without Passages} \\
\cmidrule(lr){3-5} \cmidrule(lr){6-10} \cmidrule(lr){11-15}
                                   & &EN  &ZH  &KO  & KMS    & R-2   & R-L   & BLEU  &B-Score & KMS & R-2 & R-L & BLEU &B-Score\\
\midrule
\multirow{8}{*}{\textsc{EN}}
&\textsc{Llava1.5} \cite{llava}       &\xmark &\xmark &\xmark & \textbf{88.4} & 26.0  & 37.2  & 14.7  & 76.3  & 25.6 & 17.8 & 30.4 & 9.1    & 73.4 \\
&\textsc{Llava-Llama-3}             &\xmark &\xmark &\xmark & 79.2 & 25.4  & 38.8  & 13.5  & 79.3  & 20.4  & 12.2 & 23.8 & 6.1   & 73.4 \\
&\textsc{Llava-Llama-3+VLR-IF(En)}   &\cmark &\xmark &\xmark & 85.6 & 30.1  & 46.4  & 20.9  & 81.5  & 20.4 & 19.1   & 29.9 & 8.6  & 69.1\\

&\textsc{X-Llava} \cite{shin2024xllava}                  &\xmark &\xmark &\xmark & 80.4 & 28.1  & 42.2  & 16.9  & 80.1  & 20.8  & 17.5 & 31.9 & 9.6   & 74.3 \\
&\textsc{X-Llava+VLR-IF(En)}        &\cmark &\xmark &\xmark & 82.4 & 29.4  & 44.2  & 20.1  & 80.7  & 20.4  & 19.7 & 35.4 & 12.7  & 77.5 \\
&\textsc{X-Llava+VLR-IF(En+Ko)}     &\cmark &\xmark &\cmark & 83.2 & 30.2  & 45.2  & 20.6  & 81.0  & 18.4  & 20.4 & 36.3 & 14.5  & 77.1 \\

&\textsc{Qwen-VL-Chat}              &\xmark &\xmark &\xmark & 84.8 & 32.8  & 47.4  & 20.6  & 82.1  & 31.2  & 20.0 & 34.1 & 9.7  & 77.5 \\


&\textsc{GPT-4o} \cite{openai2024gpt4}                   &\xmark &\xmark &\xmark & 85.6 & \textbf{42.6}  & \textbf{57.9}  & \textbf{32.8}  & \textbf{85.6}  & \textbf{61.6}  & \textbf{35.6} & \textbf{52.1} & \textbf{26.2}  & \textbf{83.7} \\

\midrule
\multirow{3}{*}{\textsc{ZH}}
&\textsc{QWEN-VL-CHAT} \cite{bai2023qwenvl}           &\xmark &\xmark &\xmark & 75.6 & 51.6  & 56.3  & 33.8  & 84.0 & 10.8  & 28.9 & 37.2 & 18.2  & 75.4 \\
&\textsc{QWEN-VL-CHAT+VLR-IF(Zh)} &\xmark &\cmark &\xmark & 72.4 & \textbf{59.0}  & \textbf{63.4}  & \textbf{42.9}  & \textbf{86.2} & 16.0 & 30.4 & 37.8 & 18.1  & 77.4 \\
&\textsc{GPT-4o} &\xmark &\xmark &\xmark & \textbf{80.4} & 56.9  & 62.3  & 41.6  & \textbf{86.2} & \textbf{36.0}  & \textbf{36.6} & \textbf{42.9} & \textbf{24.6}  & \textbf{80.3} \\

\midrule
\multirow{4}{*}{\textsc{KO}}
&\textsc{X-Llava}               &\xmark &\xmark &\xmark & 59.6  & 27.0 & 35.2 & 15.2  & 78.4  & 6.0 & 18.0 & 28.0 & 8.6 & 74.2 \\
&\textsc{X-Llava+VLR-IF(Ko)}     &\xmark &\xmark &\cmark & 63.6  & 35.7 & 44.1 & 24.9  & 81.0  & 6.8 & 22.4 & 32.9 & 14.9 & 77.0 \\
&\textsc{X-Llava+VLR-IF(En+Ko)}  &\cmark &\xmark &\cmark & 62.4  & 36.0 & 44.6 & 24.2  & 81.7  & 0.8 & 4.7  & 15.2 & 5.14 & 64.5 \\
&\textsc{GPT-4o} &\xmark &\xmark &\xmark & \textbf{83.6} & \textbf{51.9}  & \textbf{55.2}  & \textbf{37.2}  & \textbf{84.4} & \textbf{31.6}  & \textbf{35.9} & \textbf{39.0} & \textbf{24.9}  & \textbf{79.7} \\

\bottomrule
\end{tabular}
\caption{Overall experiment results on \textsc{VLR-Bench} depending on its language. (R: Rouge and B-Score: Bert-Score)}
\label{tab:experiment_all}
\end{table*}

\paragraph{Image Selection.} Images are crucial within this dataset. The diversity of categories among the selected images is essential for depicting a range of external knowledge. 
Considering these factors, we manually curated 150 images from BOK-VQA, developed explicitly for open-world QA purposes.

We manually extracted 150 images from the 10 categories proposed by BOK-VQA, with 15 images each from the object-centric, atmosphere-centric, and relation-centric categories. In addition, We collected 150 images of different languages' cultural backgrounds from Wikimedia Commons under the same conditions as BOK-VQA.

\paragraph{Question Selection.} The question selection process used GPT-4o to receive recommendations for high-quality question-answer pairs. We input images into GPT-4o and requested them to generate ten queries, two essential pieces of external knowledge required to resolve these queries, and descriptive answers. To ensure the validity of the model verification, we imposed the following conditions: (1) The generated data should consist of question-answer pairs that cannot be resolved with the image alone. (2) Image information should not be explicitly evident in the questions to ensure that queries cannot be resolved using external knowledge alone. 
The data produced consisted of queries related to each sample image, two pieces of external knowledge necessary to solve the queries, and a descriptive answer. At this stage, we selected the most suitable samples from the ten recommended query-knowledge pairs and conducted a preliminary review to verify that all the data consisted of queries requiring external knowledge.

\paragraph{Generation of Additional External Knowledge} 
\textsc{VLR-Bench} consists of five pieces of external knowledge. Among these, two are directly referenced 
when generating answers for the actual images and questions, referred to as `\texttt{Gold Passage}', which were already reviewed in the previous stage. Two of the five passages relate to the theme of the image or question but diverge from the central theme of the answer, termed `\texttt{Silver Passage}'. The last one, unrelated to the image and the question, is designated as `\texttt{Bronze Passage}'. At this stage, we generated two silver passages and one bronze passage. 
Three annotators directly reviewed the data derived through this process for the question-answer pairs, external knowledge, and descriptive answers. Specifically, errors in the generated external knowledge or knowledge with unclear sources were replaced with new information by annotators (see Appendix~\ref{appendix:construction_process}). Finally, each annotator extracted the two essential keywords necessary to resolve the questions. Each sample comprises five elements: an image, a query, five pieces of knowledge, a descriptive answer, and two keywords. Examples of the data are shown in Figure~\ref{fig:V-RAG data exam}.

\subsection{VLR-IF Dataset}
To address the proposed benchmark, we designed instruction-following data to enhance the utilization of external knowledge using VLMs. As previously proposed, we generated data using the same GPT-based method for question-external knowledge-answer creation. Initially, we randomly selected 9K COCO~\citep{coco} images and generated a `valid passage’ related to each image. Subsequently, we randomly extracted external knowledge from different data samples for use as `invalid passages’, thus contrastively constructing datasets using a combination of valid and invalid passages. The \textsc{VLR-IF} dataset was constructed in parallel for three languages: English, Chinese, and Korean, with each language comprising 32K data samples. The specific process for constructing the datasets is described in Appendix~\ref{vlr-if_process}.

\begin{table}[]
    \centering
    \tiny
    \begin{tabular}{lllll}
        \toprule
        LMM                    & LLM               & \#PT     & \#VIT & Language\\
        \midrule
        \textsc{Llava1.5}      & Llama2-13~B        & 558~K     & 665~K  & En,Ko,Zh\\
        \textsc{Llava-Llama-3} & Llama3-8~B         & 1.2~M     & 1.2~M  & En\\
        \textsc{X-Llava}       & Llama2-13~B        & 1.2~M     & 407~K  & En,Ko\\
        \textsc{Qwen-VL}       & Qwen 7~B           & 1.4~B     & 350~K  & Zh, En\\
        \bottomrule
    \end{tabular}
    \caption{VLMs for evaluation on \textsc{VLR-Bench}}
    \label{tab:model_desc_table}
\end{table}

\section{Experiments and Analysis}
\label{sec:Experiments}

We selected the top-performing models for each language for our experiments. Table~\ref{tab:model_desc_table} presents the base models, pre-training volumes, and visual instruction tuning (VIT) training volumes for the models used in this experiment. The \textsc{VLR-Bench} task involves generating long-form answers to the given queries. As described in  Section~\ref{sec:dataset}, two keywords were manually annotated for each query. Therefore, these keywords in the model-generated long-form answers allow for some degree of quantitative evaluation, defined as the keyword-matching score (KMS). We considered a response correct only when both answer keywords were accurately identified. 
However, because the KMS performance may improve as the generated response lengthens, it is used as a reference indicator rather than an exact performance measure. To compensate for this, a comprehensive evaluation should be conducted using metrics that account for sentence length, such as Rouge~\citep{rouge}, BLEU~\citep{bleu}, and BERT-Score~\citep{bert-score}.

Table~\ref{tab:experiment_all} presents the evaluation results for each language-specific model. If the proposed \textsc{VLR-IF} data were used for training the models, it was denoted as \textsc{+vlr-if}; the hyperparameters used in this case can be found in Appendix~\ref{appendix:hyperparameters}.

\subsection{Experiment Results}

\paragraph{Diversity in Performance Evaluation.} 
Upon examining the English KMS performance in the With Passage section of Table~\ref{tab:experiment_all}, it can be observed that the performance of \textsc{Llava1.5} closely mirrors that of GPT-4o. This raises the question of whether \textsc{Llava1.5} truly makes accurate predictions. The answer is no. The task involves generating long-form answers, and \textsc{Llava1.5} often directly outputs the received external information, resulting in lengthy responses. Although such responses achieve high KMS performance, they also contain external knowledge irrelevant to the query, leading to lower BLEU and Rouge scores.

\paragraph{The Impact of Use of External Knowledge.} 
\textsc{VLR-Bench} allows for evaluations in scenarios where external knowledge is provided, as each problem is accompanied by five pieces of external knowledge. Table~\ref{tab:experiment_all} presents the \textsc{VLR-Bench} evaluation results based on the availability of external knowledge for each model. Notably, the performance of the \textsc{X-Llava} model dropped by an average of 37.72\% for R-2 in English compared to when external knowledge was provided. These results suggest that the \textsc{VLR-Bench} dataset contains queries that require external knowledge.

\paragraph{The Impact of VLR-IF Training.}
We conducted experiments to assess the utility of the \textsc{VLR-IF} data using the baseline \textsc{Llava-Llama-3} and its version enhanced by VLR-IF training. According to the results in Table~\ref{tab:experiment_all}, the model trained with the \textsc{VLR-IF} data showed a 22.67\% performance improvement over the baseline model when external knowledge was provided. This significant enhancement suggests that the \textsc{VLR-IF} training data effectively boosts the ability to select and utilize external knowledge. Finally, we examined whether VLR-IF could positively impact other evaluation datasets, using the InfoSeek~\citep{infoseek} benchmark as a reference. The results indicated a 3.6\% performance improvement with the application of VLR-IF (see Appendix~\ref{app:deep-analysis}).

\paragraph{Comparing GPT-4o with Open Models.}

We conducted experiments to test if GPT-4o could solve \textsc{VLR-Bench} problems without external knowledge. The results from the "Without Passages" section in Table~\ref{tab:experiment_all} show that GPT-4o outperformed \textsc{qwen-vl-chat} by an average of 17.33 points without external knowledge. However, with passages provided, the performance gap narrowed to an average of 7.36 points. This indicates that \textsc{VLR-Bench} is a challenging benchmark without external knowledge, and open-source models can improve with passage-retrieval capabilities.
\subsection{Analysis}

\begin{table}[!t]
\tiny
\centering
\begin{tabular}{llcccc}
\toprule
\textbf{Lang.} &\textbf{Passage-type} & \textbf{BERT Score f1} & \textbf{Rouge-1} & \textbf{Rouge-2} & \textbf{Rouge-L} \\
\midrule[\heavyrulewidth]
\multicolumn{6}{>{\columncolor[gray]{.9}}c}{\textbf{Passages \& Ground-truth output}}\\

\midrule
\multirow{3}{*}{EN}  & Gold    & \textbf{78.04} & \textbf{44.96} & \textbf{20.98} & \textbf{31.92}  \\
                     & Silver  & 72.11 & 25.59 & 5.978 & 18.92  \\
                     & Bronze  & 67.81 & 22.10 & 2.299 & 17.04  \\
                     
\midrule
\multicolumn{6}{>{\columncolor[gray]{.9}}c}{\textbf{Passages \& Questions}}\\
\midrule
\multirow{3}{*}{EN}  & Gold    & \textbf{66.55} & \textbf{28.42} & \textbf{4.67} & \textbf{18.93}  \\
                     & Silver  & 65.70 & 21.48 & 1.97 & 15.79  \\
                     & Bronze  & 64.48 & 24.11 & 2.74 & 17.26  \\
                     
\bottomrule
\end{tabular}
\caption{Correlation analysis between Passages, Ground Truth, and Questions for the English cases.}
\label{tab:analysis_output_passage_en}
\end{table}

In this section, we present an in-depth analysis to determine whether the \textsc{VLR-Bench} is a suitable dataset for evaluating model’s ability to utilize information. To this end, we measured the BERT-score and Rouge scores (R-1, R-2, R-L) between passage types, questions, and ground-truth outputs. The results presented in Table~\ref{tab:analysis_output_passage_en} show that the ground truth output correlates most strongly with the \texttt{Gold} - \texttt{Silver} - \texttt{Bronze Passage} in descending order. This trend substantiates the effective use of gold passages in deducing answers to the \textsc{VLR-Bench}, indicating that the appropriate utilization of externally sourced knowledge through images is crucial for answering queries. On the other hand, an examination of the passages and question results reveals no clear trend, as Bronze’s Rouge-1 score is higher than Silver’s, suggesting that selecting suitable external knowledge based solely on the query can be challenging. This implies that understanding the images is necessary.
\section{Conclusion}
\label{sec:Conclusion}

In this study, we propose \textsc{VLR-Bench} for evaluating RAG-based VLMs, and \textsc{VLR-IF} for performance enhancement. The proposed benchmark differs from existing external knowledge-based VLM evaluation datasets in the following ways. (1) It consists of problems that are difficult to solve without external knowledge. (2) It includes five different passages, allowing the test of an ability not covered in previous research to determine which passages are useful for answering queries. The training data were designed as multilingual evaluation data that could simultaneously assess English, Chinese, and Korean, enhancing their utility. 
\section{Limitations}
\label{sec:Limitations}

In this study, we proposed a benchmark and corresponding training data to evaluate the RAG capabilities of VLMs. The benchmark allows for the evaluation of both retrieval and generation abilities. However, there are still two issues that remain:

\paragraph{Absence of Image Search Capability.}
Ultimately, the ability to perform image searches is crucial for accurately assessing the performance of the VLR-Bench. As mentioned in Table~\ref{tab:experiment_all}, the superior performance of GPT-4o over other public language models originates from the presence or absence of image search capabilities. Unfortunately, this study did not consider methods related to image search.

\paragraph{Lack of Diversity in Responses Due to Training Data Construction Costs.}
The method proposed in this study enabled the construction of training data at a very low cost. However, applying the same method to other languages still incurs costs, particularly when building test data, which can be expensive. Due to these cost constraints, annotation was performed by a single individual. While there could be multiple correct answers to the short-answer core keywords, due to budget limitations, responses were collected from only one person. Nevertheless, the final test data underwent a secondary review process to ensure data quality.
\section*{Acknowledgments}
This work was supported by Institute of Information \& communications Technology Planning \& Evaluation (IITP) grant, funded by the Korea government (MSIT) (No.RS-2024-00456709, A Development of Self-Evolving Deepfake Detection Technology to Prevent the Socially Malicious Use of Generative AI) and Artificial intelligence industrial convergence cluster development project funded by the Ministry of Science and ICT (MSIT, Korea)\& Gwangju Metropolitan City awarded to KyungTae Lim.
\bibliography{coling_latex}
\onecolumn
\appendix

\addtocontents{toc}{\protect\setcounter{tocdepth}{2}}

\hypersetup{linkcolor=black}
\renewcommand{\contentsname}{} 
\vspace*{0.5cm} 
\setlength{\cftaftertoctitleskip}{6em} 

\begin{center}
    {\Huge \textbf{Appendix}} 
\end{center}

\setlength{\cftbeforesecskip}{1.5em} 
\setlength{\cftbeforesubsecskip}{1em}

\tableofcontents
\clearpage

\section{VLR-Bench}
\label{appendix:vlr-bench}
\subsection{VLR-Bench Examples}
\noindent \textbf{Overall} The following figures are examples from \textsc{VLR-Bench}. Each example consists of a question, an answer, keywords, and passages. The ``\texttt{gold passage}'', which contains the information necessary to answer the question, is highlighted in yellow.
\begin{figure}[!h]
    \centering
    \includegraphics[width=\textwidth]{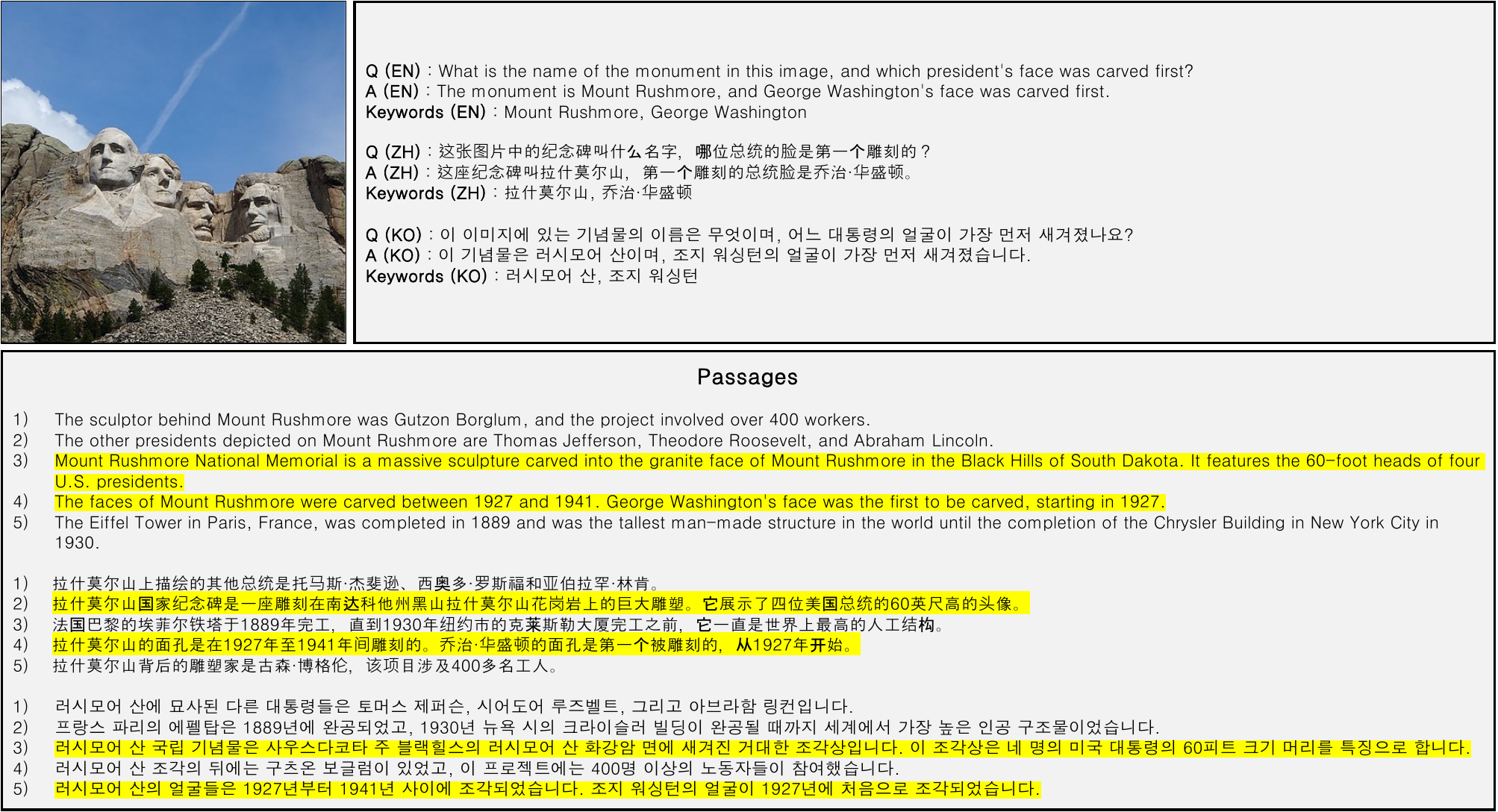}
    \caption{Examples of the created VLR-Bench data. (English culture)}
    \label{fig:enter-label}
\end{figure}
\begin{figure}[!h]
    \centering
    \includegraphics[width=\textwidth]{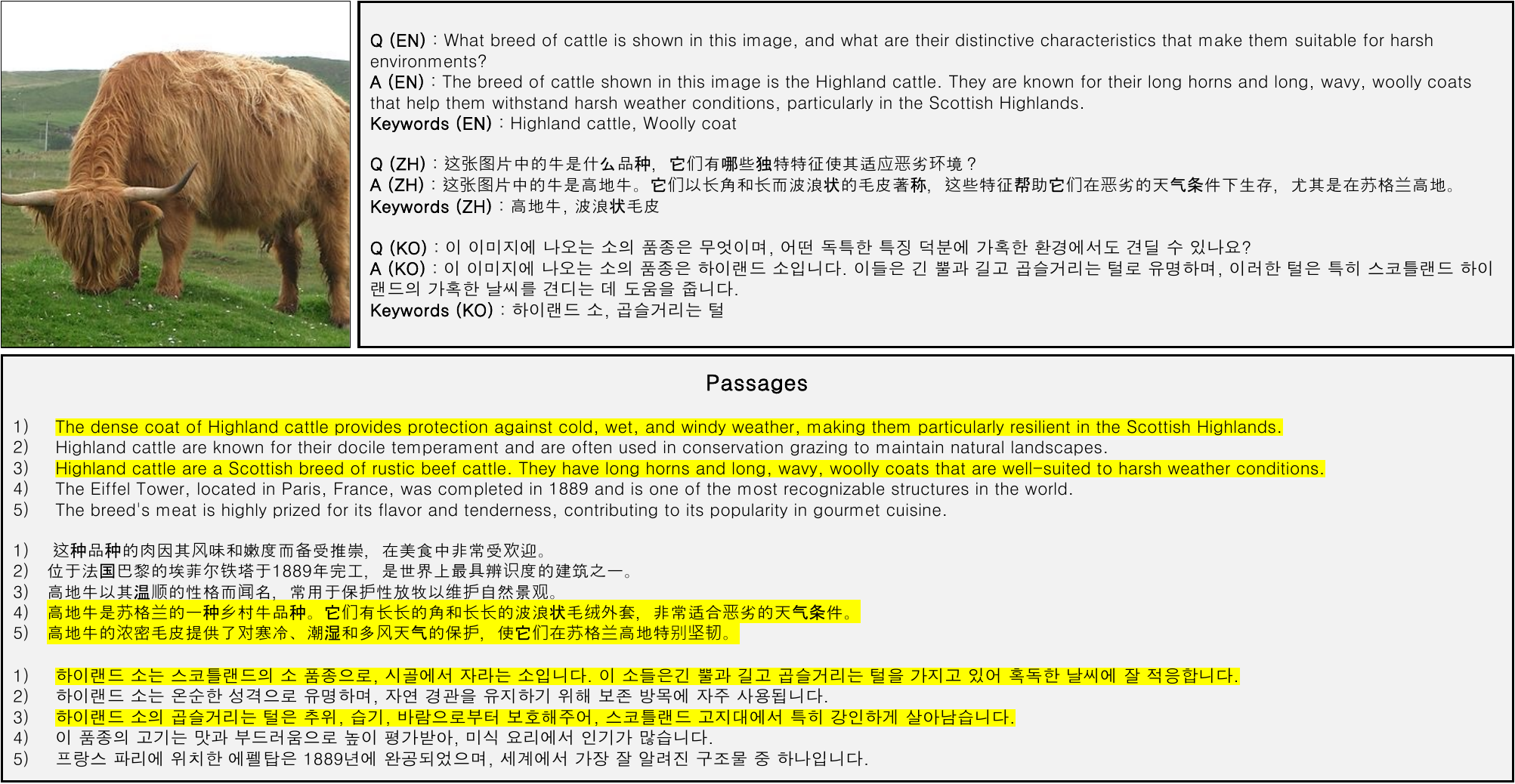}
    \caption{Examples of the created VLR-Bench data. (commonsense knowledge)}
    \label{fig:enter-label}
\end{figure}

\noindent \textbf{Data incorporating language-specific cultural aspects.} \textsc{VLR-Bench} comprises 150 datasets for each language, incorporating language-specific cultural aspects. The benchmark is designed to include queries that require an understanding of the respective culture to accurately select the correct information from the provided passages. Without the requisite cultural knowledge, identifying the appropriate passage becomes challenging, even when given access to the entire set of passages.
\begin{figure}[!h]
    \centering
    \includegraphics[width=\textwidth]{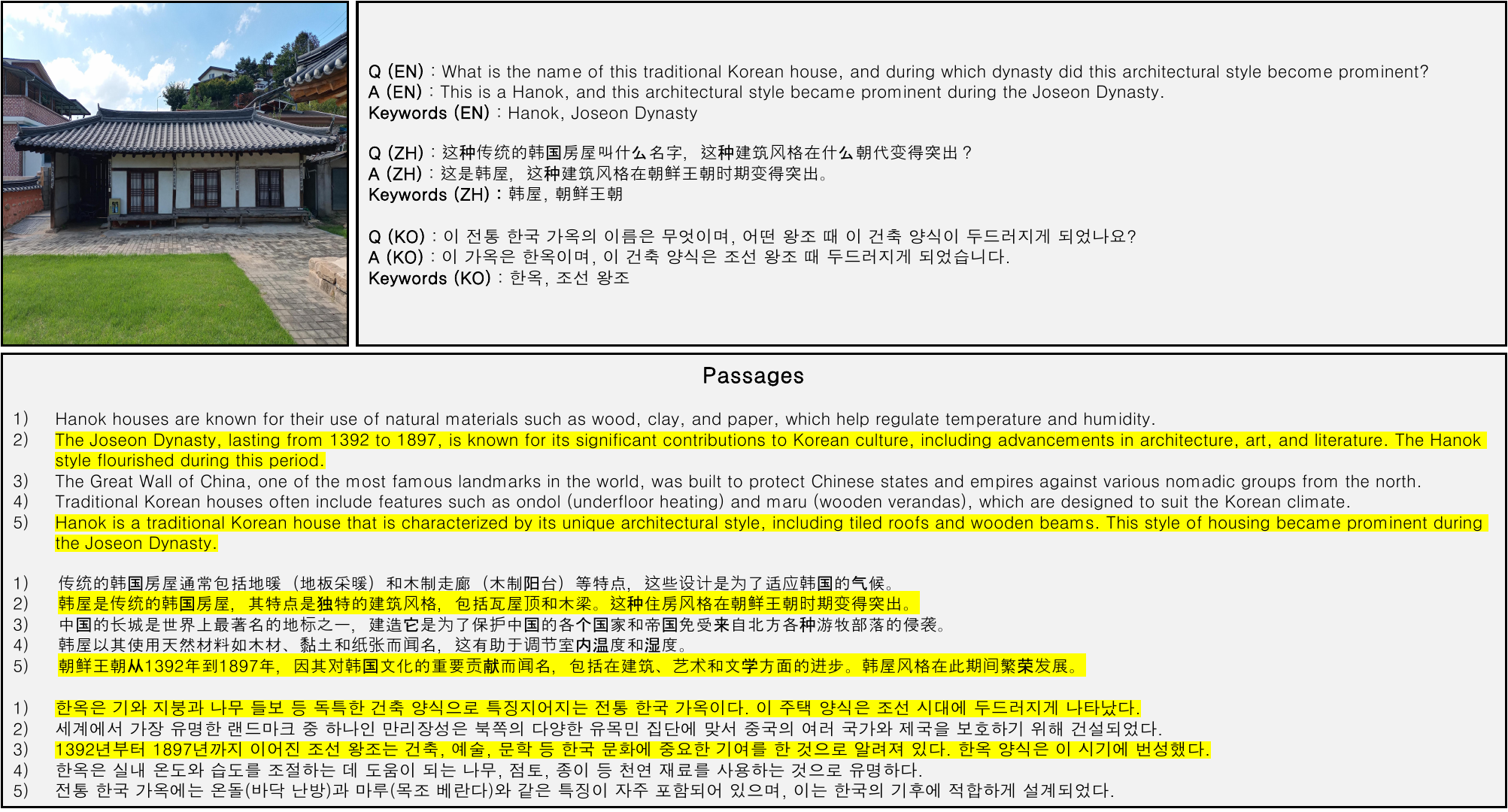}
    \caption{Examples of the created VLR-Bench data. (Korean culture)}
    \label{fig:enter-label}
\end{figure}
\begin{figure}[!h]
    \centering
    \includegraphics[width=\textwidth]{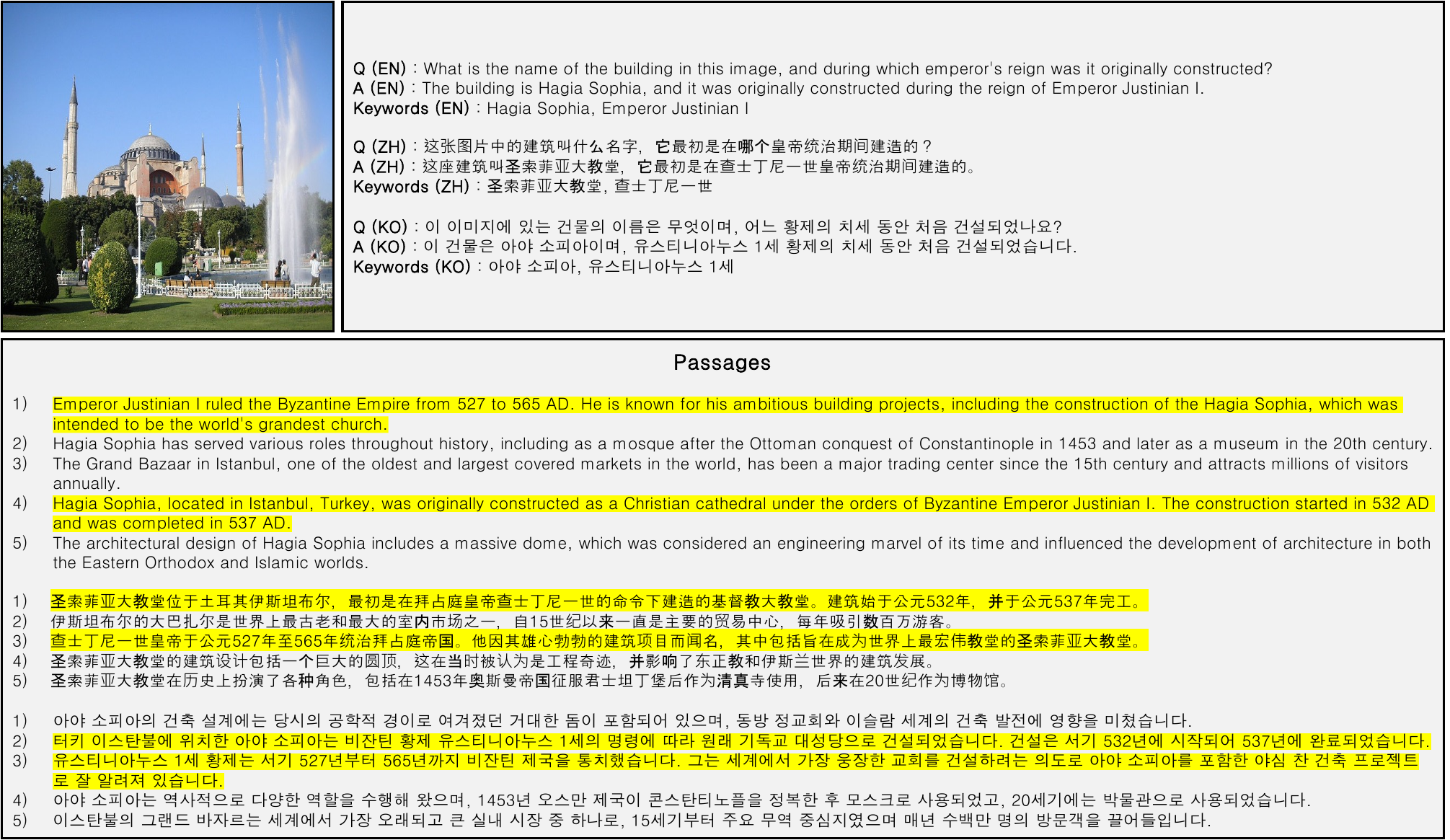}
    \caption{Examples of the created VLR-Bench data. (commonsense knowledge)}
    \label{fig:enter-label}
\end{figure}

\clearpage
\subsection{\textsc{VLR-Bench} Construction Process}\label{appendix:construction_process}
\begin{figure}[!h]
    \centering
    \includegraphics[width=\textwidth]{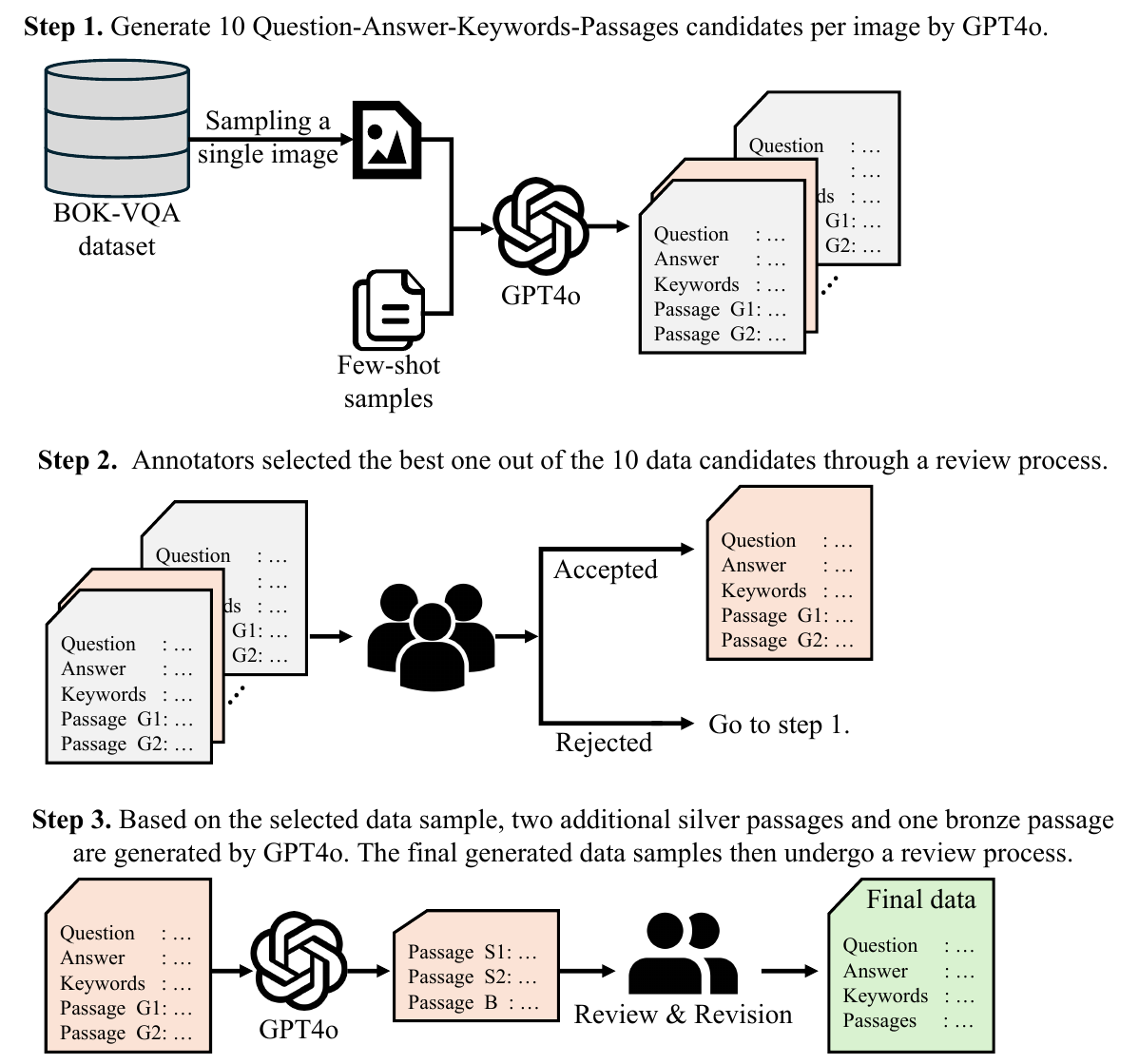}
    \caption{Overview of the \textsc{VLR-Bench} dataset construction process.}
\end{figure}
\paragraph{Overview of Data Construction Procedure} The image samples used in the dataset are sourced from the BOK-VQA~\citep{kim2024bokvqa} dataset, ensuring a wide range of visual content. The construction process involves few-shot learning and initial generation, annotator review and selection, passage expansion, and final review. GPT-4o generates candidate question-answer-passage sets based on few-shot examples, which are then reviewed and selected by human annotators. The selected sets are further expanded by GPT4o to create additional silver and bronze passages. The final dataset comprises a query, an answer, five passages (two gold, two silver, and one bronze), and two answer keywords for each image. Through a rigorous review process, the dataset maintains a high level of quality and relevance.

\paragraph{Annotation Guidelines}

To ensure the production and verification of high-quality data, we employed three computer science students. The annotators, aged 23, 23, and 27, included native speakers of Chinese and Korean, who were responsible for data in their respective languages. To generate data optimized for model training, we adhered to the guidelines for long-form sentences provided by BOK-VQA. However, when determining the Gold, Silver, and Bronze status of external knowledge, which is not covered by BOK-VQA, the annotators used their personal judgment. We proposed a maximum sentence length of 200 tokens for external knowledge. In cases where there were discrepancies in the corrections among annotators, discussions were held to revise in a more natural direction. Specifically, during the final data construction, there were many conflicts in selecting two keywords depending on the annotator’s preferences. Therefore, a 27-year-old annotator proficient in both Chinese and Korean made the final selection by choosing two keywords from all the ones that had been selected at least once.

\begin{figure}[!h]
    \centering
    \includegraphics[width=\textwidth]{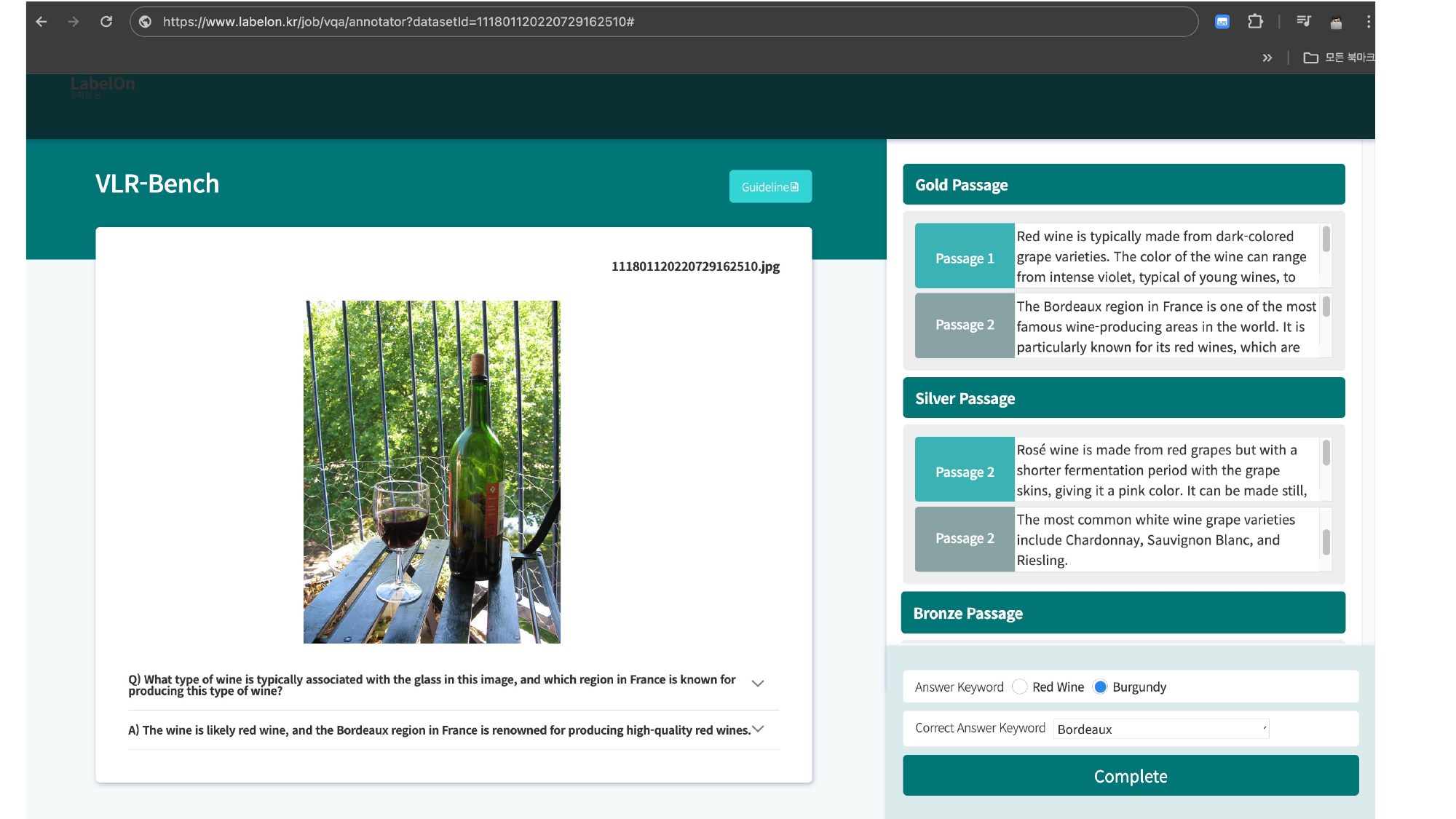}
    \caption{\textsc{VLR-Bench} annotation tool.}
\end{figure}

\clearpage
\subsection{\textsc{VLR-Bench} Few-shot Setup Examples}
As mentioned in Subsection~\ref{appendix:construction_process}, we generate 10 question-answer-keywords-passages candidates using few-shot samples. In this section, we demonstrate our few-shot examples.
\begin{figure}[!h]
  \includegraphics[width=\textwidth]{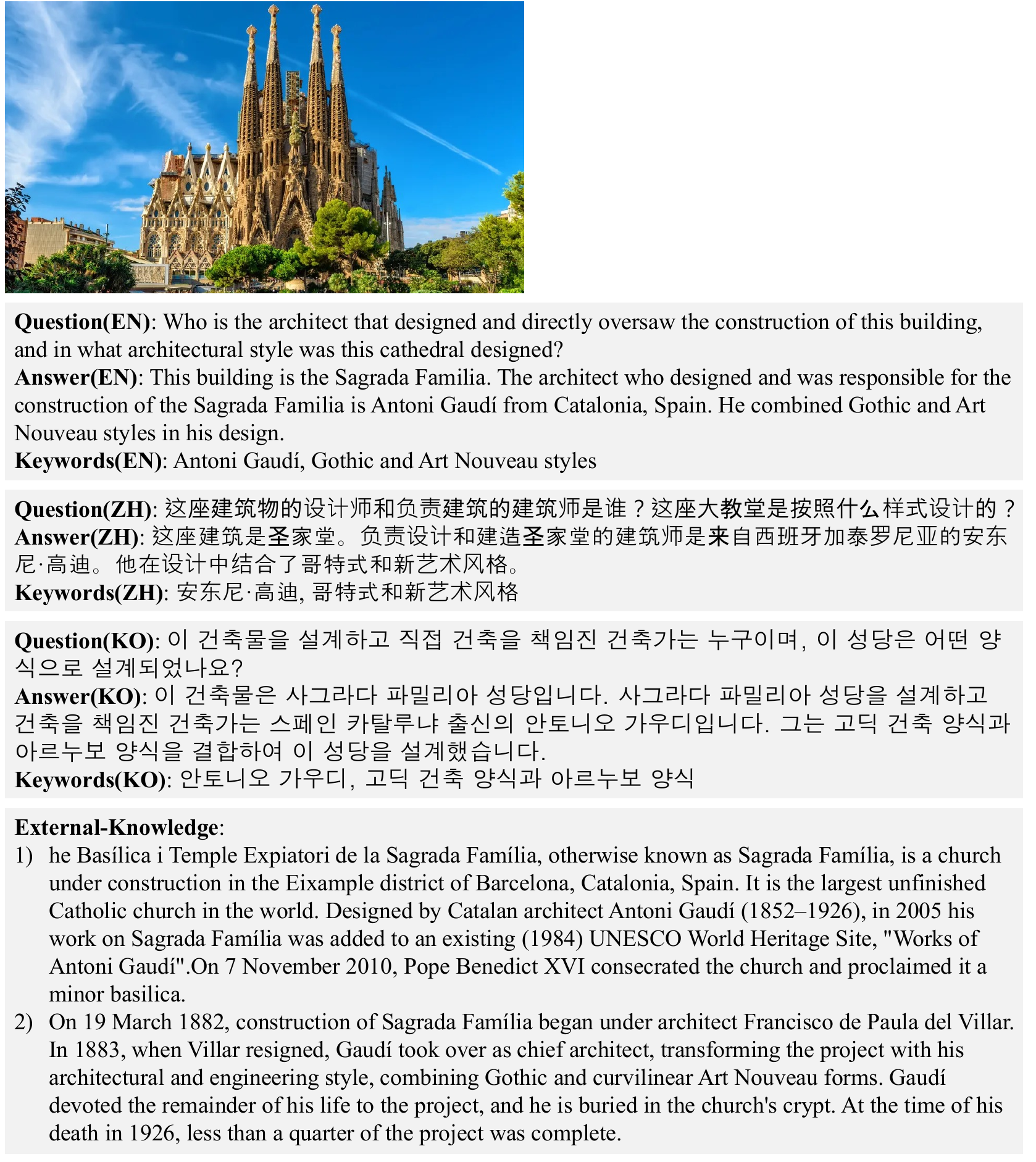}
  \caption{Examples of the few-shot sample.}
  \label{fig:vlr-if example-1}
\end{figure}

\label{sec:Analysis}

\clearpage
\newpage 

\section{VLR-IF}
\subsection{\textsc{VLR-IF} Example}
The following figures are examples from \textsc{VLR-IF}. Each example consists of a question, an answer, and a passage.

\begin{figure}[!h]
  \includegraphics[width=\linewidth]{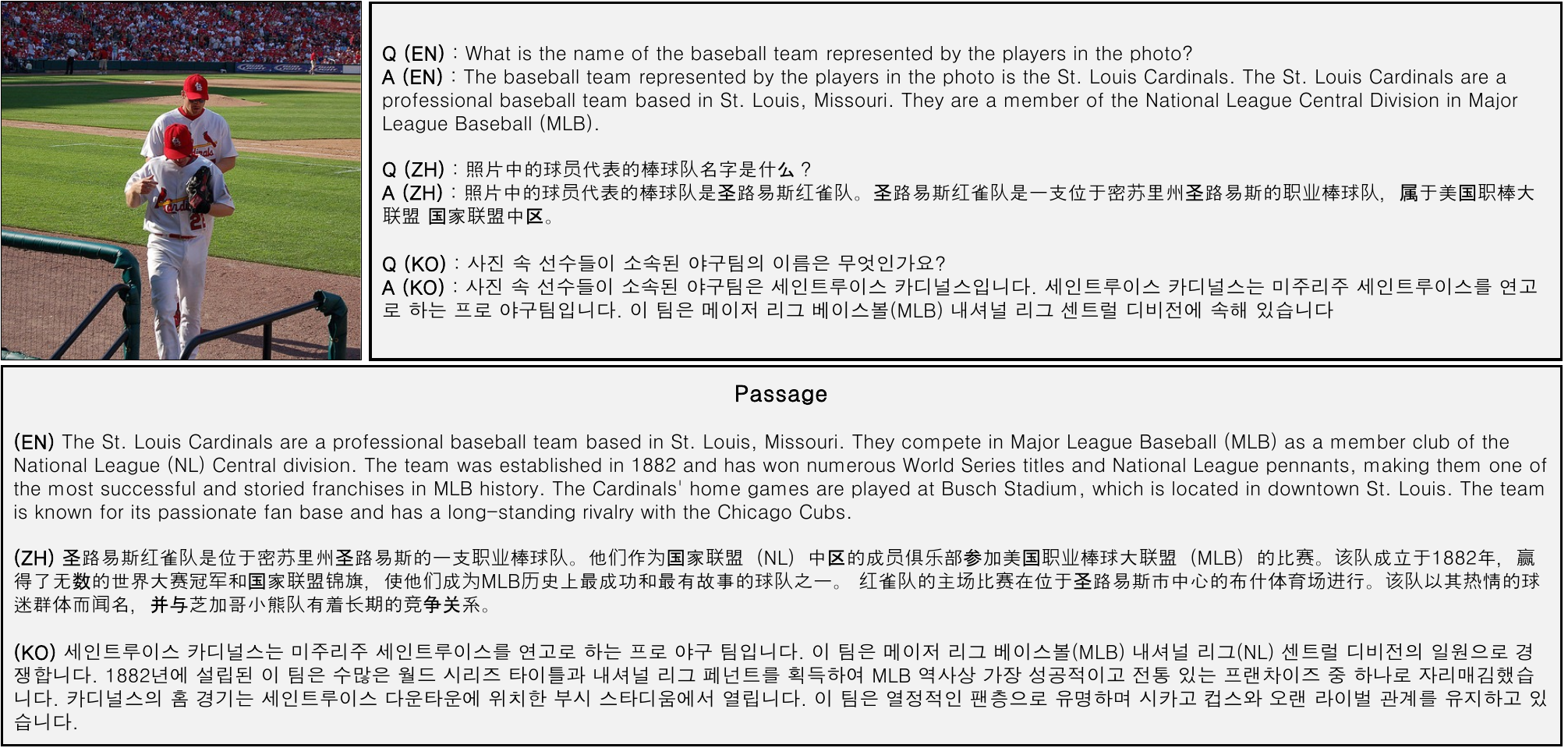}
  \caption{First example of the created VLR-IF data.}
  \label{fig:vlr-if example-1}
\end{figure}

\begin{figure}[!h]
  \includegraphics[width=\linewidth]{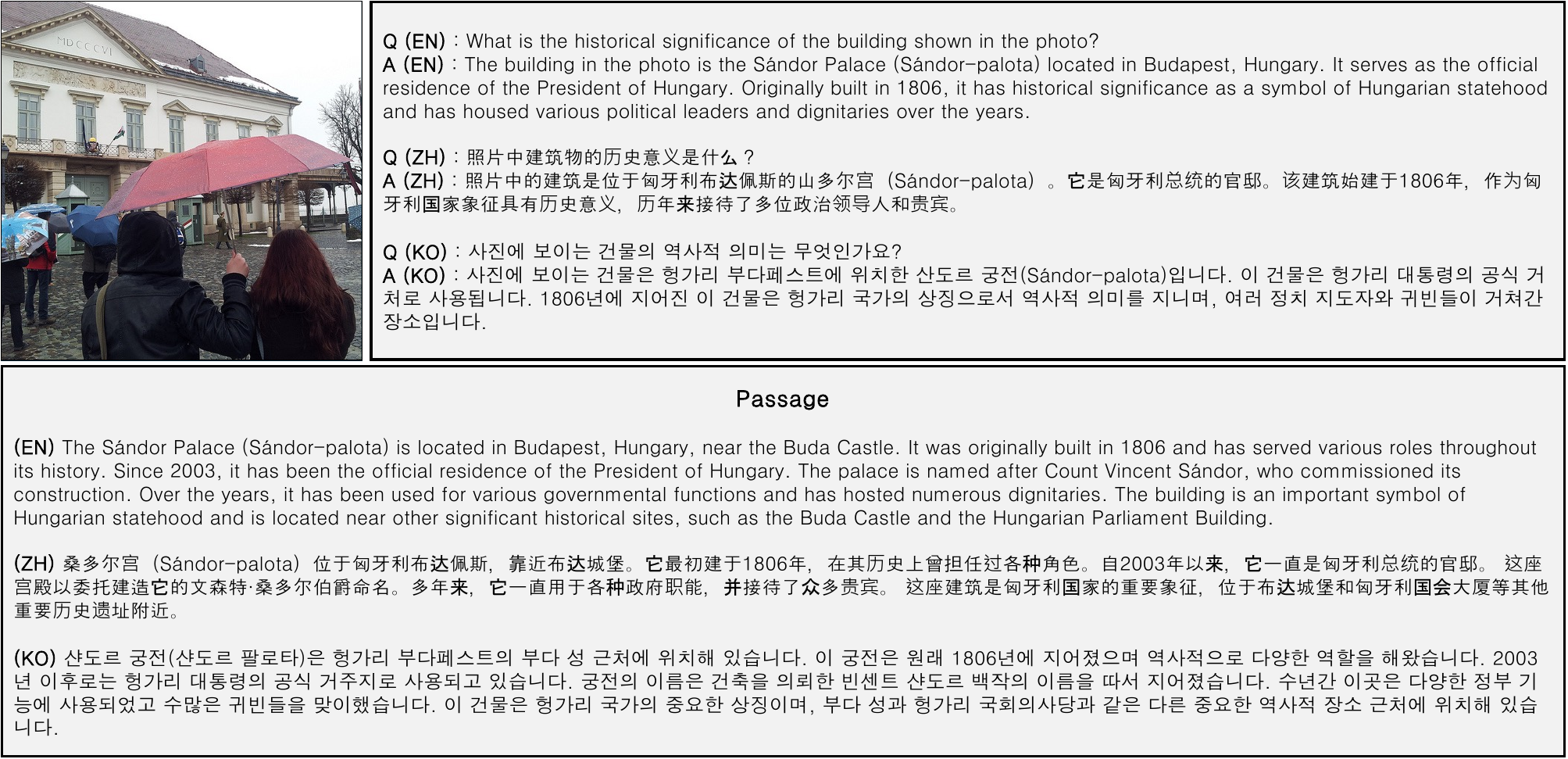}
  \caption{Second example of the created VLR-IF data.}
  \label{fig:vlr-if example-2}
\end{figure}

\subsection{\textsc{VLR-IF} Construction Process}
\label{vlr-if_process}
\begin{figure}[h]
  \includegraphics[width=\linewidth]{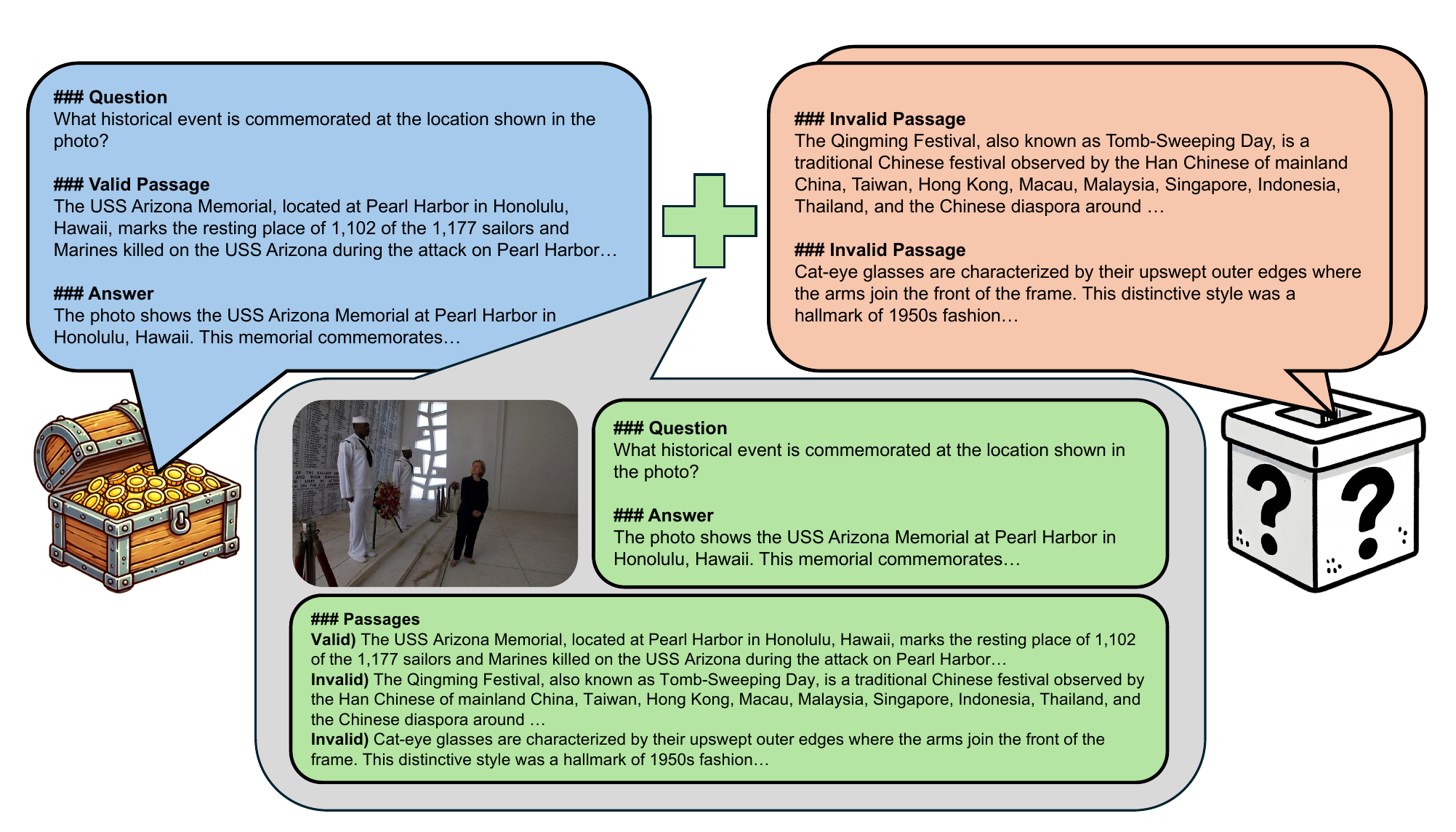}
  \caption{The process of constructing the VLR-IF dataset.}
  \label{fig:vlr-if process}
\end{figure}

Figure~\ref{fig:vlr-if process} illustrates the construction process of the \textsc{VLR-IF} dataset. The dataset consists of 9K images, with each image corresponding to a single query, answer, and passage. Following the approach used in building VLR-Bench, we provided GPT-4o with few-shot samples (including image, query, answer, and passage) along with the image example we wanted to generate. Then, we generated the query, answer, and passage for the image example. To enhance the model's ability to select valid passages, we determined that it would be desirable to include diverse passages for each image. Accordingly, we assumed the original passage of each image to be a valid passage and randomly extracted passages from other images to set them as invalid passages. When only invalid passages are used, the model is designed to generate the following response: ``The provided knowledge does not pertain to the image, so I can't answer the question.'' Ultimately, we constructed a total of 32,000 datasets by combining valid and invalid passages in the following manner: $\{V\}$, $\{I\}$, $\{V, I\}$, $\{V, I, I\}$.

\begin{itemize}
    \item $\{V\}$: Only the valid passage, 9,000 datasets.
    \item $\{I\}$: Only one invalid passage, 5,000 datasets. In this case, the training data was constructed to output "Insufficient search results found, making inference impossible" when encountering such instances.
    \item $\{V, I\}$: One valid and one invalid passage, 9,000 datasets.
    \item $\{V, I, I\}$: One valid and two invalid passages, 9,000 datasets.
\end{itemize}

\section{Details of Experimental Environments}
\label{appendix:hyperparameters}

\subsection{Baseline Models}
\begin{table}[h]
    \centering
    \small
    \begin{tabular}{llllr}
        \toprule
        LMM                    & LLM         & \#VIT & source & latest update(dd.mm.yyyy)\\
        \midrule
        \textsc{Llava1.5}      & Llama2-13~B & 665~K & liuhaotian/llava-v1.5-13b                 & 10.05.2024 \\
        \textsc{Llava-Llama-3} & Llama3-8~B  & 1.2~M & xtuner/llava-llama-3-8b-v1\_1-transformers & 28.04.2024\\
        \textsc{X-Llava}       & Llama2-13~B & 407~K & MLP-KTLim/X-LLaVA                           & 02.01.2024\\
        \textsc{Qwen-VL}       & Qwen 7~B    & 350~K & Qwen/Qwen-VL-Chat                           & 26.01.2024\\
        \bottomrule
    \end{tabular}
    \caption{The Vision-Language Models (VLMs) used for evaluation on \textsc{VLR-Bench} were accessed through the Hugging Face Transformers library version 4.32.0~\cite{wolf-etal-2020-transformers}}
    \label{tab:model_source_table}
\end{table}

\paragraph{\textsc{Llava-Llama-3}.} The LLaVA-based model fine-tuned from meta-llama/Meta-Llama-3-8B-Instruct\footnote{https://huggingface.co/meta-llama/Meta-Llama-3-8B-Instruct} and CLIP-ViT-Large-patch14-336\footnote{https://huggingface.co/openai/clip-vit-large-patch14-336} with ShareGPT4V-PT and InternVL-SFT by XTuner.
\paragraph{\textsc{X-Llava}.}
We selected X-LLaVA, a Korean and English Multimodal LLM, as the base model for the Korean and English benchmarks. X-LLaVA was trained on a dataset of 91K English-Korean-Chinese multilingual and multimodal learning data.
\paragraph{\textsc{Qwen-VL-Chat}.}
We employed Qwen-7B as the LLM and Openclip ViT-bigG as the Visual Encoder. The Qwen-VL model is constructed by connecting the LLM and Visual Encoder to a randomly initialized cross-attention layer. Finally, Qwen-VL-Chat is a model obtained by fine-tuning Qwen-VL using an instruction-following dataset.

\subsection{Hyperparameter Settings}
\begin{table}[!ht]
\centering
{\footnotesize
    \begin{tabular}{l | c } 
    \hline
{} & {value} \\ 
 \hline
 Optimizer & AdamW \\ 
 learning\_rate & 5.0e-5  \\
 Dropout & 0.05 \\ 
 lr\_scheduler & cosine \\
 Epoch for IT & 1  \\ 
 Epoch for PT & 1  \\ 
 sequence\_len & 4096 \\
 Batch size   & 1  \\
  Random Seed & 1004  \\ 
 \hline
 llm & lora \\
 Low-rank size & 64  \\
 lora\_alpha & 128 \\
 lora\_dropout & 0.05 \\
 lora\_trainable & q, v, k, o, gate, down, up\_proj \\
 LoRA layer & q, k, v \\
 \hline
 
  \end{tabular}
}
\caption{Applied hyperparameter settings.}   
\label{tab:hyperparameter}
\end{table}

The hyperparameter settings used in this study can be found in Table~\ref{tab:hyperparameter}. Models utilizing LoRA were trained using only a portion of the attention layers indicated in the table, as well as $\theta^{e}$ and $\theta^{h}$, and the size of the low-rank matrices was set to 64. All models were trained for 1 epoch.

\paragraph{Experiment Reproduction.}
We are making the training code, trained models, and data used for testing available to allow for exact reproduction of the experiments conducted in this study. The qualitative responses generated by the models during the experiments can be downloaded from the following site, with files named after the models corresponding to the experimental results of those models.

\subsection{Performance Comparison on Various Passage Types.}
\label{app:deep-analysis}
\begin{table}[!h]
    \small
    \centering
    \begin{tabular}{llccccc}
    \toprule
    \multirow{2}{*}{Model} & \multicolumn{6}{c}{English} \\
    \cmidrule(lr){2-7} 
    & PSG & MS & R-2 & R-L & BLEU & BERT-Score \\
    \midrule
    \textsc{LLaVA-LLaMA-3}                  & GG  & \textbf{84.8} & \textbf{39.9} & \textbf{50.2} & \textbf{20.6} & \textbf{84.1} \\
                                            & G   & 58.4 & 30.9 & 41.4 & 14.8 & 81.2 \\
                                            & GS  & 68.0 & 33.5 & 44.3 & 16.2 & 82.0 \\
                                            & GB  & 59.6 & 30.4 & 40.9 & 14.6 & 81.0 \\
                                            & SS  & 41.2 & 23.3 & 33.7 & 10.6 & 78.3 \\
                                            & SB  & 38.4 & 23.4 & 33.5 & 10.4 & 78.3 \\
                                            & B   & 22.4 & 19.5 & 29.6 & 9.2  & 76.4 \\
    \midrule
    \textsc{LLaVA-LLaMA-3+VLR-IF}           & GG & \textbf{86.4} & \textbf{49.5} & \textbf{62.4} & \textbf{34.0} & \textbf{87.3} \\
                                            & G  & 68.0 & 42.6 & 56.1 & 26.1 & 85.1 \\
                                            & GS & 73.6 & 42.1 & 55.9 & 26.2 & 85.0 \\
                                            & GB & 69.2 & 41.6 & 55.3 & 26.9 & 84.8 \\
                                            & SS & 48.0 & 33.3 & 47.1 & 18.5 & 81.9 \\
                                            & SB & 46.0 & 33.9 & 47.9 & 19.6 & 82.1 \\
                                            & B  & 22.4 & 22.0 & 35.0 & 15.0 & 76.7 \\
    \bottomrule
    \end{tabular}    
    \caption{Performance Comparison of LLaVA-LLaMA-3 with and without VLR-IF Training on Various Passage Types. The results demonstrate that, regardless of the passage type, the model trained on VLR-IF consistently outperforms its counterpart without VLR-IF training across all evaluation metrics. This finding supports the hypothesis that the VLR-IF dataset effectively enhances the model's ability to select crucial information from passages, enabling it to better follow user instructions based on the given image.}
    \label{tab:performance_wrt_passage}
\end{table}

\begin{table}[h]
    \small
    \centering
    \begin{tabular}{lc}
    \toprule
    \multicolumn{1}{c}{Model} & \multicolumn{1}{c} {\textsc{InfoSeek}}\\
    \midrule
    \textsc{Llava-Llama-3}                  & 42.9   \\
    \textsc{Llava-Llama-3+VLR-IF(EN)}       & 44.5   \\
    \bottomrule
    \end{tabular}
    \caption{Performance difference in InfoSeek depending on VLR-IF training when using a search engine as a passage retriever.}
    \label{tab:infoseek}
\end{table}

Table~\ref{tab:infoseek} presents the results of evaluating the InfoSeek benchmark performance with and without VLR-IF training using the \textsc{llava-llama-3} model. VLR-IF was trained solely on the English dataset, and the Oracle was used as the Retriever model. The evaluation showed that the model trained with VLR-IF achieved a 2.6 points improvement in performance even on the external benchmark dataset, InfoSeek.

\section{Comprehensive Analysis of Datasets}

\subsection{\textsc{VLR-Bench} Validity Analysis}

\begin{table*}[!h]
\tiny
\centering
\begin{tabular}{llccccc}
\toprule
\multirow{2}{*}{Lang.} & \multirow{2}{*}{Model} &\multicolumn{3}{c}{VLR-IF} & \multirow{2}{*}{Quantitative Avg.} & \multirow{2}{*}{GPT4o Score} \\
\cmidrule(lr){3-5} 
                                   & &EN  &ZH  &KO  & \\
\midrule
\multirow{8}{*}{\textsc{EN}}
&\textsc{Llava1.5} \cite{llava}       &\xmark &\xmark &\xmark  & 48.5 & 9.11 \\
&\textsc{Llava-Llama-3}             &\xmark &\xmark &\xmark  & 48.2 & 9.03 \\
&\textsc{Llava-Llama-3+VLR-IF(En)}   &\cmark &\xmark &\xmark  & 52.9 & \textbf{9.40}\\

&\textsc{X-Llava} \cite{shin2024xllava} &\xmark &\xmark &\xmark  & 49.5 & 9.10\\
&\textsc{X-Llava+VLR-IF(En)}        &\cmark &\xmark &\xmark  & 51.4 & 9.27\\
&\textsc{X-Llava+VLR-IF(En+Ko)}     &\cmark &\xmark &\cmark  & 52.1 & 9.28\\
&\textsc{Qwen-VL-Chat}              &\xmark &\xmark &\xmark  & \textbf{53.5} & 9.30\\

\midrule
\multirow{3}{*}{\textsc{ZH}}
&\textsc{QWEN-VL-CHAT} \cite{bai2023qwenvl} &\xmark &\xmark &\xmark  &60.3 & 8.33\\
&\textsc{QWEN-VL-CHAT+VLR-IF(Zh)} &\xmark &\cmark &\xmark  &\textbf{64.8} & \textbf{9.26}\\

\midrule
\multirow{3}{*}{\textsc{KO}}
&\textsc{X-Llava}               &\xmark &\xmark &\xmark   & 43.1& 7.62 \\
&\textsc{X-Llava+VLR-IF(Ko)}     &\xmark &\xmark &\cmark  & \textbf{50.0}& 8.35\\
&\textsc{X-Llava+VLR-IF(En+Ko)}  &\cmark &\xmark &\cmark  & 49.8& \textbf{8.59}\\

\bottomrule
\end{tabular}
\caption{A table illustrates the results of the qualitative assessment using GPT-4o. Quantitative Avg. is the average result of the quantitative evaluation conducted Table~\ref{tab:experiment_all}.}
\label{tab:quan_qual_table}
\end{table*}

To validate the quantitative evaluation results of \textsc{VLR-Bench}, we conducted a qualitative assessment using GPT-4o. GPT-4o was provided with images from the \textsc{VLR-Bench} dataset, queries, external knowledge required for answering, and the model's responses. Based on this information, the model's responses were evaluated on the following four aspects:
(1) Assessment of the model's selection of Gold Passages and the use of Silver Passages.
(2) Evaluation of the accuracy, completeness, and readability of the model's responses.
(3) Verification of the model's fulfillment of the query requirements.
(4) Examination of whether additional content from Silver Passages or Bronze Passages was included based on the length of the responses.
This rigorous evaluation ensures the reliability and validity of the quantitative results obtained from \textsc{VLR-Bench}.

\begin{figure}[!h]
 \includegraphics[width=\linewidth]{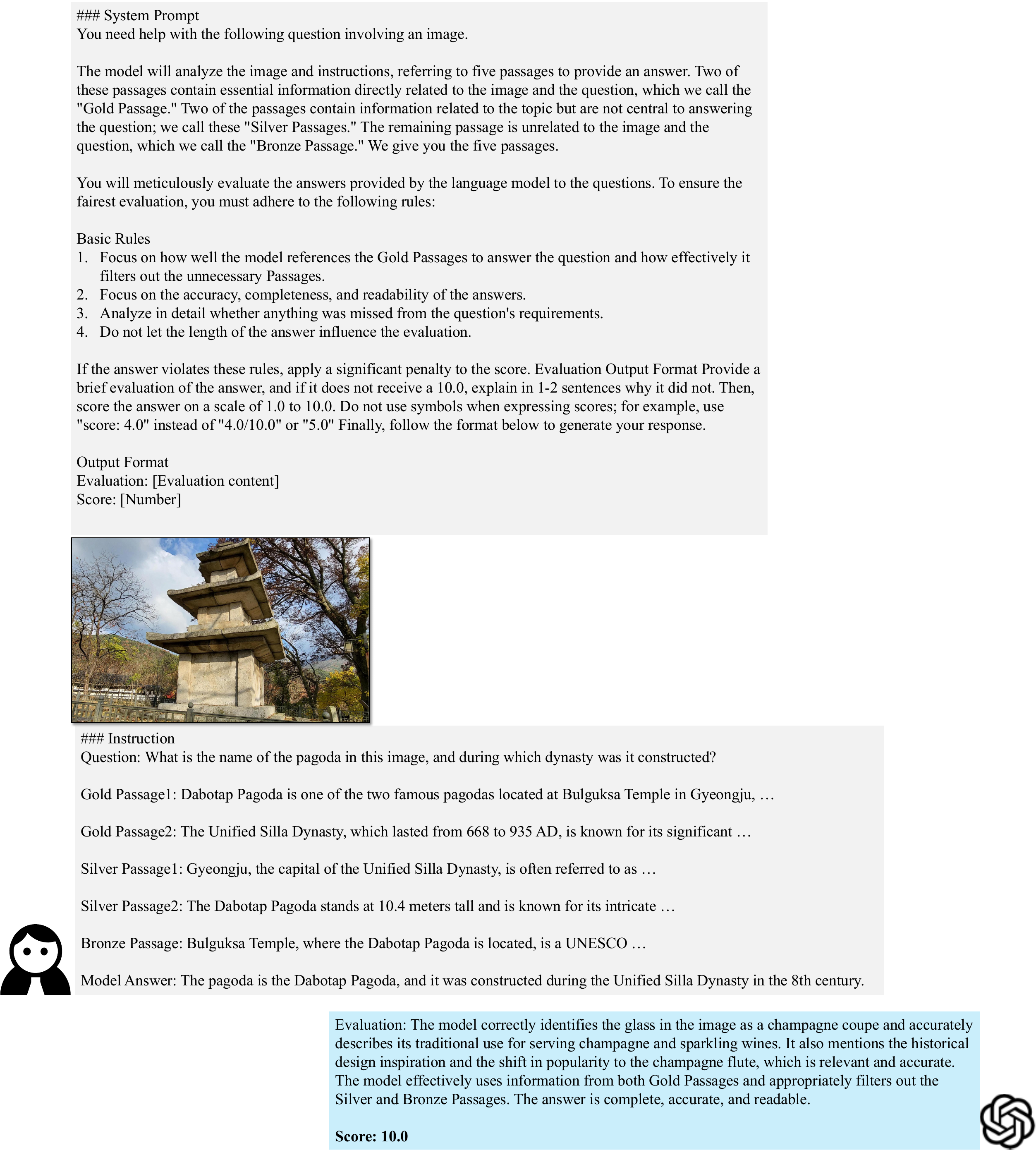}
 \captionsetup{font=small}
 \caption{Examples of prompts used with GPT models in qualitative evaluations.}
 \label{fig:qualitative-eval-prompt}
\end{figure}

Additionally, GPT-4o outputs the evaluation scores along with the reasoning behind the evaluations. Through this process, we conducted a reliable qualitative assessment and confirmed that the results exhibited a distribution similar to the quantitative evaluation results of VLR-Bench, as shown in Table~\ref{tab:quan_qual_table}. The Figure~\ref{fig:qualitative-eval-prompt} illustrates the prompt and responses provided to GPT-4o for the qualitative assessment.

\subsection{\textsc{VLR-IF} Validity Analysis}
To investigate the impact of the VLR-IF dataset, we evaluated its effect on passage selection in our experiments. We chose English as the target language for the experiments and used the Llava-Llama-3 model, which received the highest evaluation in this language. The experiment proceeds as follows: first, we provide the model with passages, an instruction, and an image. The model then selects two passages necessary to follow the instruction related to the image. As shown in the Table~\ref{tab:Choice_score}, the model fine-tuned on VLR-IF demonstrates a substantial improvement of 26.0 and 25.1 points in EM and F1 scores, respectively, compared to the model without VLR-IF training. The results suggest that the VLR-IF dataset can enhance the ability to select the necessary passages based on images and queries.

\begin{table}[h]
    \small
    \centering
    \begin{tabular}{lcc}
    \toprule
    \multicolumn{1}{c}{Model} & \multicolumn{1}{c}{EM} & \multicolumn{1}{c}{F1}\\
    \midrule
    \textsc{Llava-Llama-3}                  & 2.0  & 15.9 \\
    \textsc{Llava-Llama-3+VLR-IF(EN)}       & 28.0 & 41.0  \\
    \bottomrule
    \end{tabular}
    \caption{Passage Selection Performance with and without VLR-IF Training. EM is the exact matching score, while F1 is the harmonic mean of precision and recall.}
    \label{tab:Choice_score}
\end{table}

\clearpage
\newpage

\subsection{Related Datasets}
\label{sec:appendix}

\begin{table*}[ht!]
    \centering
    \small 
    \begin{adjustbox}{max width=\textwidth}
        \begin{tabular}{llrcccccc}
            \toprule
                Dataset & Image Source     & \# of Instances & Multilingual & Parallel & Open   & Qualitative & Quantitative  & With Passages\\
                \midrule
                  K-VQA &     Wikipedia         &   183~K   & \xmark       & \xmark   & \cmark & \xmark      & \cmark  & \xmark \\
                  S3VQA & Open Images Dataset   &     6~K   & \xmark       & \xmark   & \cmark & \xmark      & \cmark  & \xmark \\
                 OK-VQA &        COCO           &    14~K   & \xmark       & \xmark   & \cmark & \xmark      & \cmark  & \xmark \\
                A-OKVQA &       COCO            &    24~K   & \xmark       & \xmark   & \cmark & \xmark      & \cmark  & \xmark \\
                ViQuAE &     Wikipedia          &     3~K   & \xmark       & \xmark   & \cmark & \xmark      & \cmark  & \xmark \\
              OVEN-Wiki &     Wikipedia         &   5,8~M   & \xmark       & \xmark   & \cmark & \xmark      & \cmark  & \xmark \\
               InfoSeek &     Wikipedia         &  1.36~M   & \xmark       & \xmark   & \cmark & \xmark      & \cmark  & \xmark \\
       Encyclopedic VQA & iNaturalist, Google Landmarks Dataset & 1.036~M & \xmark & \xmark & \cmark & \xmark & \cmark & \xmark \\
               \rowcolor[gray]{.9}
                   Ours &        BOK-VQA          &       32.3~K      & \cmark       & \cmark   & \cmark & \cmark      & \cmark  & \cmark \\
            \hline
        \end{tabular}
    \end{adjustbox}
    \caption{Summary of the multimodal VQA (Visual Question Answering) benchmark dataset. `Parallel' indicates that the dataset can be used for translation tasks. `Qualitative' refers to the availability for quantitative evaluation, while `Quantitative' refers to the availability for qualitative evaluation. `With Passages' denotes whether passages are provided in the benchmark dataset.}
    \label{tab:dataset_compare}
\end{table*}




Table~\ref{tab:dataset_compare} provides information on the size and domains of major VLM evaluation datasets that utilize external knowledge. The \textsc{VLR-Bench} dataset proposed in this study is structurally similar to the Encyclopedic VQA dataset, which includes test data containing 1,000 gold passages. However, \textsc{VLR-Bench} differs in two key ways: (1) instead of a single gold passage, each query is paired with five passages—two \texttt{Gold Passages}, two \texttt{Silver Passages}, and one \texttt{Bronze Passage}, and (2) it consists of parallel corpora in English, Chinese, and Korean, making the test data more than four times larger, even though the total number of samples is smaller. Moreover, unlike the automatically generated Encyclopedic VQA, all passages in \textsc{VLR-Bench} have been manually reviewed, with a strong emphasis on quality control. By categorizing passages into gold, silver, and bronze, models must distinguish between useful and less relevant information to generate accurate answers. This design allows for a more nuanced evaluation of how well a VLM can utilize gold passages while avoiding the silver and bronze ones from the top-k retrieved results, setting \textsc{VLR-Bench} apart from existing datasets.

\subsection{Correlation analysis between Passages, Ground Truth, and Questions.}
\begin{table}[!h]
\tiny
\centering
\begin{tabular}{llcccc}
\toprule
\textbf{Lang.} &\textbf{Passage-type} & \textbf{Bert Score f1} & \textbf{Rouge-1} & \textbf{Rouge-2} & \textbf{Rouge-L} \\
\midrule[\heavyrulewidth]
\multicolumn{6}{>{\columncolor[gray]{.9}}c}{\textbf{Passages \& Ground-truth output}}\\
\midrule
\multirow{3}{*}{EN}  & Gold    & \textbf{78.04} & \textbf{44.96} & \textbf{20.98} & \textbf{31.92}  \\
                     & Silver  & 72.11 & 25.59 & 5.978 & 18.92  \\
                     & Bronze  & 67.81 & 22.10 & 2.299 & 17.04  \\
                     
\midrule
\multirow{3}{*}{ZH}  & Gold    & \textbf{77.08} & \textbf{20.78} & \textbf{7.16} & \textbf{20.50}  \\
                     & Silver  & 70.49 & 4.15  & 1.08  & 4.15  \\
                     & Bronze  & 66.28 & 0.38  & 0.0   & 0.38   \\
                     
\midrule
\multirow{3}{*}{KO}  & Gold    & \textbf{77.20} & \textbf{19.05} & \textbf{6.90} & \textbf{18.78}  \\
                     & Silver  & 71.98 & 3.47  & 0.77 & 3.47  \\
                     & Bronze  & 66.48 & 0.38  & 0.0  & 0.38  \\
\midrule
\multicolumn{6}{>{\columncolor[gray]{.9}}c}{\textbf{Passages \& Questions}}\\
\midrule
\multirow{3}{*}{EN}  & Gold    & \textbf{66.55} & \textbf{28.42} & \textbf{4.67} & \textbf{18.93}  \\
                     & Silver  & 65.70 & 21.48 & 1.97 & 15.79  \\
                     & Bronze  & 64.48 & 24.11 & 2.74 & 17.26  \\
                     
\midrule
\multirow{3}{*}{ZH}  & Gold    & \textbf{67.13} & \textbf{1.2}   & 0.0  & \textbf{1.2}  \\
                     & Silver  & 65.85 & 0.2   & 0.0  & 0.2  \\
                     & Bronze  & 63.70 & 0.0   & 0.0  & 0.0  \\
                     
\midrule
\multirow{3}{*}{KO}  & Gold   & \textbf{68.15} & \textbf{1.2}   & 0.0  & \textbf{1.2}   \\
                     & Silver & 67.26 & 0.2   & 0.0  & 0.2   \\
                     & Bronze & 66.28 & 0.380 & 0.0  & 0.380  \\

\bottomrule
\end{tabular}

\caption{Examining the correlation between Passages and GT reveals that, irrespective of the language used, the correlations are ordered in the sequence of Gold, Silver, and Bronze. This suggests that to successfully perform \textsc{VLR-Bench}, it is necessary to appropriately utilize the Gold Passage. Meanwhile, investigating the correlation between Passages and Questions indicates that the level of correlation remains consistent across various types of Passages. These results demonstrate that Questions alone are insufficient for successfully completing \textsc{VLR-Bench}, and that both images and Passages must be utilized together.}

\label{tab:analysis_output_passage}
\end{table}

\end{document}